\documentclass{article}
\usepackage{spconf,amsmath,graphicx,hyperref}
\usepackage{amssymb}
\usepackage{booktabs}   % 支持 \toprule 等
\usepackage{multirow}   % 支持 \multirow
\usepackage{multicol}   % 支持 \multicolumn
\usepackage{tabularx}   % 支持自动列宽
\usepackage{hyphenat}

\title{Physics-Driven 3D Gaussian Rendering for Zero-Shot MRI Super-Resolution}
\name{Shuting Liu$^{1}$ \qquad Lei Zhang$^{1,*}$ \qquad Wei Huang$^{1,*}$ \qquad Zhao Zhang$^{1}$ \qquad Zizhou Wang$^{2}$\thanks{$*$ Corresponding author. This work was supported by the Fundamental and Interdisciplinary Disciplines Breakthrough Plan of the Ministry of Education of China (JYB2025XDXM101) and Sichuan Province Innovative Talent Funding Project for Postdoctoral Fellows(BX202512).}}
\address{$^{1}$School of Artificial Intelligence, Sichuan University, Chengdu 610065, China\\$^{2}$Institute of High Performance Computing, A*STAR, Singapore 138632}
\begin{document}
\ninept
\maketitle
%
% \begin{abstract}
% The abstract should appear at the top of the left-hand column of text, about
% 0.5 inch (12 mm) below the title area and no more than 3.125 inches (80 mm) in
% length.  Leave a 0.5 inch (12 mm) space between the end of the abstract and the
% beginning of the main text.  The abstract should contain about 100 to 150
% words, and should be identical to the abstract text submitted electronically
% along with the paper cover sheet.  All manuscripts must be in English, printed
% in black ink.
% \end{abstract}
% %
% \begin{keywords}
% One, two, three, four, five
% \end{keywords}
%

\begin{abstract}
High-resolution Magnetic Resonance Imaging (MRI) is vital for clinical diagnosis but limited by long acquisition times and motion artifacts. Super-resolution (SR) reconstructs low-resolution scans into high-resolution images, yet existing methods are mutually constrained: paired-data methods achieve efficiency only by relying on costly aligned datasets, while implicit neural representation approaches avoid such data needs at the expense of heavy computation. We propose a zero-shot MRI SR framework using explicit Gaussian representation to balance data requirements and efficiency. MRI-tailored Gaussian parameters embed tissue physical properties, reducing learnable parameters while preserving MR signal fidelity. A physics-grounded volume rendering strategy models MRI signal formation via normalized Gaussian aggregation. Additionally, a brick-based order-independent rasterization scheme enables highly parallel 3D computation, lowering training and inference costs. Experiments on two public MRI datasets show superior reconstruction quality and efficiency, demonstrating the method's potential for clinical MRI SR.
\end{abstract}


\section{Introduction}
\label{sec:intro}
Magnetic resonance imaging (MRI) provides superior soft-tissue contrast and is indispensable for many clinical tasks\cite{huang2022BANet, huang2022LANet}, but acquiring high-resolution (HR) volumes requires prolonged scan time and stronger gradients\cite{jia2017new,wu2022arssr}, which increase safety risks and motion artifacts. Consequently, clinical protocols often favor faster~\cite{wang2018physicsMRI}, lower-resolution (LR) acquisitions that suffer from partial-volume effects and reduced anatomical fidelity. Super-resolution (SR) reconstruction aims to recover HR details from such efficiently acquired LR scans, alleviating this trade-off.

Early MRI SR methods relied on interpolation~\cite{keys1981cubic} and model-based optimization~\cite{yang2010sr_sparse}. With deep learning, approaches from SRCNN to EDSR and RCAN enabled end‑to‑end LR–HR mapping~\cite{dong2014srcnn,lim2017edsr,zhang2018rcan}, while transformer-based models further improved anatomical consistency~\cite{huang2025gapmatch, huang2024exploring}. More recent coordinate-based and implicit approaches inspired by NeRF/LIIF learn continuous mappings from spatial coordinates and enable arbitrary-scale SR without paired training data~\cite{mildenhall2020nerf,chen2021liif}. 
% Despite these advances, two key challenges remain. First, paired LR–HR training data are scarce in routine clinical practice because HR acquisitions prolong exam time and increase motion/SAR risk~\cite{wang2018physicsMRI}. Second, implicit zero-shot methods avoid paired data requirements but suffer from heavy computational costs, as their per-scan optimization times are prohibitively long for clinical use.
Despite these progress, two challenges remain. First, paired LR–HR scans are scarce: acquiring HR volumes requires longer scans and stronger gradients, which increase  patient discomfort and motion artifacts. Multi-site and inter-protocol variability further induce domain shifts that impair supervised models~\cite{chitnis2017variability,liu2020domainAdaptation}. Second, implicit zero-shot approaches remove the need for paired data but are computationally intensive: training a coordinate-based model per 3D volume needs dense sampling and repeated inference, resulting in per-scan optimization times of several hours~\cite{tancik2021dnerf,jagatap2022fastNeRF}.

To address both issues, we propose a physics-driven zero-shot MRI SR framework built upon an explicit 3D Gaussian point cloud. First, we design MRI-tailored Gaussian parameters that embed intrinsic tissue properties into each Gaussian, reducing learnable parameters while preserving fidelity to the MR signal equation. Second, we introduce a physics-grounded volume rendering strategy that reconstructs continuous voxel intensities through normalized aggregation of Gaussian contributions, eliminating depth sorting and view-dependent splatting. Finally, we develop a brick-based order-independent rasterizer that supports highly parallel 3D volumetric rendering via optimized CUDA kernels, substantially lowering computational costs during both training and inference. Experiments on two public MRI datasets demonstrate that our approach achieves superior reconstruction quality and efficiency, effectively addressing key challenges in zero-shot MRI super-resolution.

% Our main contributions are: 1) an explicit 3D Gaussian point-cloud model for zero-shot MRI super-resolution that yields a fully differentiable, physics-driven pipeline; 2) MRI-tailored Gaussian parameters together with a physics-grounded, normalized aggregation rendering strategy that reduces parameters while preserving MR signal fidelity; and 3) a brick-based order-independent rasterizer enabling efficient parallel 3D computation, with extensive experiments demonstrating improved accuracy and runtime on two public datasets.

We summarize our contributions as follows: 1) To the best of our knowledge, this work represents the first application of explicit 3D Gaussian point‑cloud model for zero‑shot super‑resolution of MRI data; 2) We propose MRI-Tailored Gaussian Parameters, Physics-Grounded Volume Rendering, and Brick-Based Order-Independent Rasterizer, which collectively enable high-quality 3D MRI super-resolution effectively; and 3) Extensive experiments on two public datasets demonstrate consistent and significant improvements over existing methods.

% The main contributions are summarized:
% \begin{itemize}
%   \item To the best of our knowledge, this work represents the first application of explicit 3D Gaussian point‑cloud model for zero‑shot super‑resolution of MRI data.
%   \item We propose MRI-Tailored Gaussian Parameters, Physics-Grounded Volume Rendering, and Brick-Based Order-Independent Rasterizer, which collectively enable high-quality 3D MRI super-resolution effectively.
%   % without view-dependent operations or depth sorting.
%   \item Extensive experiments on two public datasets demonstrate consistent and significant improvements over existing methods.
% \end{itemize}

\begin{figure*}[htbp]
  \centering
  \includegraphics[width=1\linewidth]{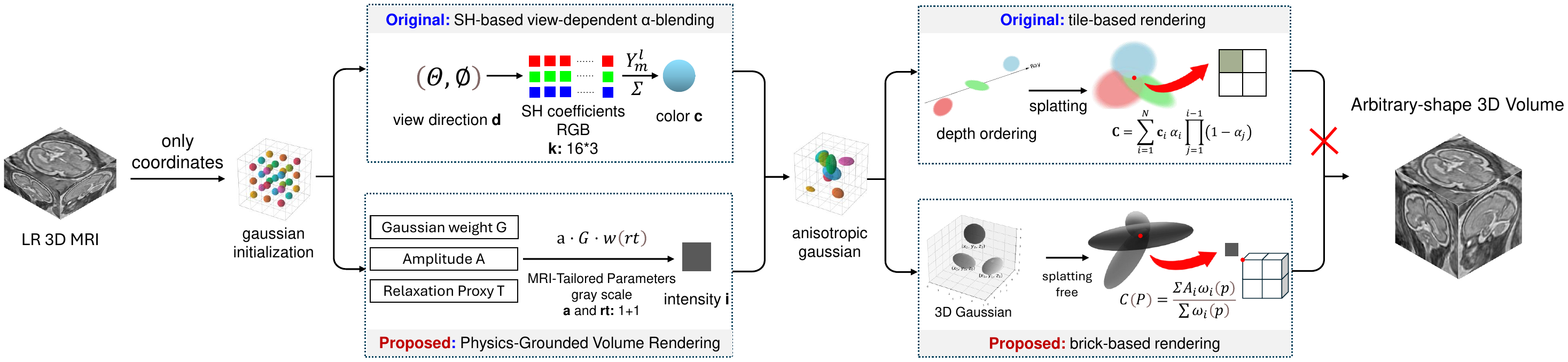}
  \caption{The overall workflow of our method. 3D Gaussians are initialized solely from the coordinates of the low-resolution volume. During rendering, 3DGS relies on spherical harmonics (SH)-based view-dependent \(\alpha\)-blending to produce RGB color, whereas our approach uses physics-grounded volume rendering to render voxel intensity. In the spatial distribution stage, 3DGS requires depth sorting followed by 2D splatting, while our method aggregates Gaussian contributions and performs brick-based rendering directly in 3D space to generate high-resolution MRI.}
  \label{fig:flowchart}
\end{figure*}

\begin{figure}[!t]
  \centering
  \includegraphics[width=\linewidth]{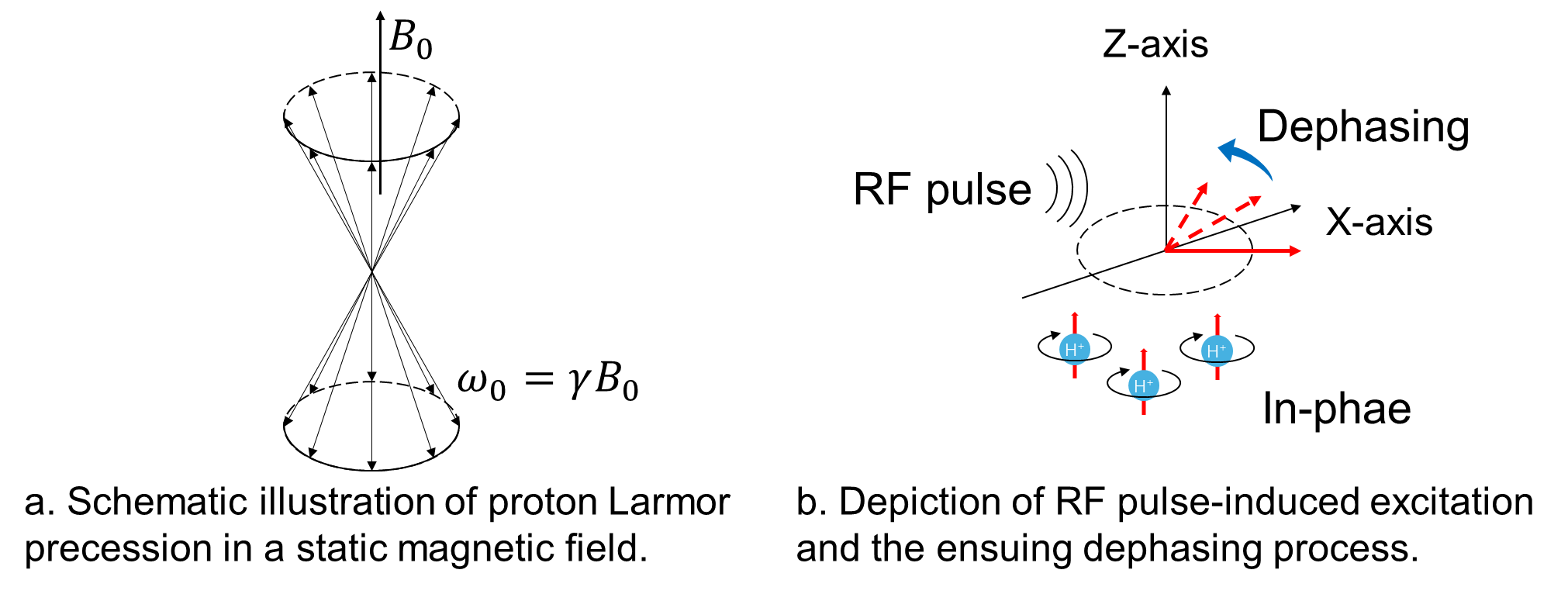}
  \caption{Illustration of the key MRI signal formation processes that challenge the original 3DGS framework and motivate our MRI-tailored Gaussian model. In MRI, protons first align with and precess in the static magnetic field $B_0$ (a), then an RF pulse tips them into the transverse plane where they dephase and relax (b), producing the time‐dependent signal used to reconstruct the image.}
  \label{fig:MRI_Principle}
\end{figure}

% \section{Formatting your paper}
% \label{sec:format}

% All printed material, including text, illustrations, and charts, must be kept
% within a print area of 7 inches (178 mm) wide by 9 inches (229 mm) high. Do
% not write or print anything outside the print area. The top margin must be 1
% inch (25 mm), except for the title page, and the left margin must be 0.75 inch
% (19 mm).  All {\it text} must be in a two-column format. Columns are to be 3.39
% inches (86 mm) wide, with a 0.24 inch (6 mm) space between them. Text must be
% fully justified.

\section{Preliminaries and Motivation}
\label{sec:preliminaries}

MRI voxel intensities arise from local magnetization induced by hydrogen protons, determined primarily by proton density $\rho(\mathbf{x})$ and longitudinal recovery. For a spatial location $p$, the observed intensity can be expressed as
\begin{equation}
\label{eq:mri_continuous}
I(p) = \int_{\mathbb{R}^3} \rho(x)\,\bigl(1 - e^{-TR/T(x)}\bigr)\,\mathrm{PSF}(p - x)\,\mathrm{d}x,
\end{equation}
where \(TR\) is the repetition time. This physical process makes MRI fundamentally view‑independent: intensities arise from hydrogen protons aligning in the static magnetic field \(B_0\), excitation by RF pulses, and transverse magnetization decay (Figure~\ref{fig:MRI_Principle}).

Directly applying 3D Gaussian Splatting (3DGS)\cite{kerbl20233dgs} to MRI creates three key mismatches with 3DGS: 1) 3DGS's spherical harmonics model view-dependent appearance, while MRI has no view-dependent lighting effects. 2) The 2D projection pipeline in 3DGS is misaligned with MRI's volumetric signal formation. 3) MRI intensities encode biophysical tissue properties rather than surface reflectance.

To bridge this gap, we reformulate the 3DGS framework in a physics‑driven manner. First, by replacing spherical harmonics and opacity with MRI‑Tailored Gaussian Parameters, we directly model MRI voxel signals rather than view‑dependent optical properties. Second, we replace traditional 2D splatting and depth sorting with a Physics‑Grounded Volume Rendering process that algorithmically simulates the MRI signal formation procedure. Third, to further enhance efficiency, we design a Brick‑Based Order‑Independent Rasterizer that enables parallel 3D MRI volumetric rendering. Together, these strategies yield an efficient, physics-driven pipeline for zero‑shot MRI super‑resolution. The overall workflow of our method is illustrated in Figure~\ref{fig:flowchart}.

% \section{PAGE TITLE SECTION}
% \label{sec:pagestyle}

% The paper title (on the first page) should begin 1.38 inches (35 mm) from the
% top edge of the page, centered, completely capitalized, and in Times 14-point,
% boldface type.  The authors' name(s) and affiliation(s) appear below the title
% in capital and lower case letters.  Papers with multiple authors and
% affiliations may require two or more lines for this information. Please note
% that papers should not be submitted blind; include the authors' names on the
% PDF.

\section{Method}

\subsection{MRI-Tailored Gaussian Parameters}
\label{sec:mtgp}
To adapt the 3D Gaussian Splatting framework to MRI signal formation, we replace the original appearance-related parameters, opacity \(\alpha_i\) and spherical harmonics coefficients \(\mathbf{c}_i^{\mathrm{SH}}\), with physically meaningful quantities that better reflect tissue properties\cite{kerbl20233dgs}. 

Specifically, we introduce two intrinsic parameters per Gaussian: the \textbf{amplitude} $A$, corresponding to local proton density ($A \approx \rho$), and the \textbf{relaxation proxy} $T$, modeling the proton relaxation behavior that governs the return of nuclear magnetization to equilibrium.

We model the effect of tissue relaxation using the physically interpretable product form
\begin{equation}\label{eq:relaxation_modulation}
T = \left(1 - e^{-TR/T_1}\right) \, e^{-TE/T_2},
\end{equation}
where $TR$ and $TE$ are the repetition and echo times, and $T_1$, $T_2$ denote the longitudinal and transverse relaxation times, respectively. The first term models longitudinal recovery after excitation, while the second term accounts for transverse decay. In our framework, we do not explicitly store $T_1$ and $T_2$ for each Gaussian. Instead, we introduce a single \textbf{relaxation proxy} $T$ that models the overall proton relaxation behavior governing the return of nuclear magnetization to equilibrium. 
% The physical quantities $T_1$ and $T_2$ appear only in the interpretation of Eq.~\eqref{eq:relaxation_modulation} or in constructing an analytical prior mapping $g(t)$, allowing the same formulation to be applied across different MRI contrasts.

By parameterizing each Gaussian with both the \textbf{amplitude} $A$ and the \textbf{relaxation proxy} $T$, our representation encodes not only spatial location and shape but also material-specific signal characteristics, yielding a more interpretable and physically grounded model. This physics-inspired parameterization also brings substantial efficiency gains: whereas the original 3DGS requires \textbf{59} parameters per Gaussian (position:3, covariance:7, SH coefficients:48, opacity:1), our formulation uses only \textbf{12} (position:3, covariance:7, amplitude:1, relaxation proxy:1). This $\sim$5$\times$ reduction preserves the core Gaussian structure while aligning the representation with MRI physics, alleviating key computational bottlenecks in volumetric medical imaging.

\subsection{Physics-Grounded Volume Rendering}
\label{sec:pgvr}
We introduce Physics-Grounded Volume Rendering, a splatting\hyp{}free rendering method that reconstructs voxel intensities by directly aggregating contributions from nearby Gaussians, eliminating the need for view-dependent projection or depth-sorted splatting. This aligns with the continuous, view-independent nature of MRI signal formation, where each voxel’s intensity reflects a localized integration of underlying tissue properties.

The reconstructed intensity at spatial location \(p\) is computed as a normalized weighted average:
\begin{equation}
\label{eq:rendering}
I(p) = \frac{\sum_{i=1}^N A_i\,w_i(p)}{\sum_{i=1}^N w_i(p)},
\end{equation}
where \(N\) is the number of Gaussians and each spatial weight is defined as
\begin{equation}
w_i(p) = \exp\!\Bigl(-\tfrac12(p-\mu_i)^\top\Sigma_i^{-1}(p-\mu_i)\Bigr)r_i,
\end{equation}
with $\mu_i \in \mathbb{R}^3$ the Gaussian center and $\Sigma_i \in \mathbb{R}^{3\times3}$ its covariance.  

Here, the Gaussian kernel captures spatial proximity, while the relaxation factor \(r_i\) models tissue-dependent signal modulation through the effective relaxation proxy \(T\). The amplitude \(A_i\) encodes proton density, determining the intrinsic signal strength contributed by each tissue component. Thus, \(A_i\) defines signal magnitude, while \(w_i(p)\) determines its spatial influence.

This formulation mirrors the MRI signal model in Eq.~\eqref{eq:mri_continuous}, where voxel intensities emerge from localized integration of neighboring tissue properties. Since all terms are differentiable, the rendering is compatible with end-to-end optimization from volumetric MRI data. To compensate for the fact that a finite Gaussian set does not form a strict partition of unity, we normalize by the accumulated weights \(\sum_i w_i(p)\). This ensures energy conservation and signal linearity, keeping the reconstruction consistent with the continuous integral formulation.

\subsection{Brick-Based Order-Independent Rasterizer}
\label{sec:bboir}
The original 3DGS framework uses a 2D tile-based splatting strategy \cite{kerbl20233dgs} for perspective rendering. For volumetric MRI we fundamentally redesign that with a 3D brick-based pipeline that respects spatial continuity, enables efficient parallelism. Our method eliminates computationally expensive sorting operations while maintaining full differentiability through optimized gradient computation.

We partition the volume into uniform bricks of size $8\times8\times4$ voxels, each handled by one CUDA thread block. For each brick $b$ we identify intersecting Gaussians $\mathcal{G}_b$ via their $3\sigma$ Mahalanobis ellipsoids, load their compact parameters into shared memory, and compute per-voxel intensities using the normalized blending equation. Because the blending is commutative and order-independent, no global depth sorting is required.

During the forward pass we cache per-voxel numerator $S(p_i)=\sum_{i}A_i\,w_i(p_i)$ and denominator $W(p_i)=\sum_{i}w_i(p_i)$. Given loss $\mathcal{L}$, the gradient w.r.t.\ spatial parameters $\theta_i\in\{\mu_i,\Sigma_i\}$ is
\begin{equation}
\frac{\partial \mathcal{L}}{\partial \theta_i}
= \sum_{p_i \in b} \frac{\partial \mathcal{L}}{\partial I(p_i)}
\left[\frac{A_i - I(p_i)}{W(p_i)}\,\frac{\partial w_i(p_i)}{\partial \theta_i}\right],
\end{equation}
and the gradient w.r.t.\ amplitudes is
\begin{equation}
\frac{\partial \mathcal{L}}{\partial A_i}
= \sum_{p_i \in b} \frac{\partial \mathcal{L}}{\partial I(p_i)}\,\frac{w_i(p_i)}{W(p_i)}.
\end{equation}
Caching $S$ and $W$ avoids re-summing over $\mathcal{G}_b$ and makes backpropagation locality-friendly.

All core operations, including brick intersection tests, shared-memory parameter loading, intensity computation, and gradient backpropagation, are implemented as optimized CUDA kernels. By eliminating global sorting and exploiting commutative volumetric blending, the rasterizer achieves substantial speedups over 2D approaches. The backward pass uses constant extra memory (two scalars per gaussian) rather than dynamic per-pixel primitive lists, and the brick-to-thread-block mapping yields natural 3D parallelism for efficient reconstruction of MRI volumes.

\begin{table}[!t]
    \centering
    \caption{PSNR/SSIM for multi-scale SR on the MSD dataset~\cite{simpson2019msd}. 
    \textbf{Bold} and \underline{underline} indicate the best and second best results.}
    \label{tab:brats_results}
    \resizebox{\linewidth}{!}{
    \begin{tabular}{lcccc}
        \toprule
        Method & 2$\times$ & 3$\times$ & 4$\times$ & Arbitrary \\
        \midrule
        \multicolumn{5}{c}{\textbf{Conventional methods}} \\
        \midrule
        Bicubic
         & 33.75, 0.947
         & 30.74, 0.916
         & 28.67, 0.872
         & 31.56, 0.933 \\
        \midrule
        \multicolumn{5}{c}{\textbf{Supervised methods}} \\
        \midrule
        MIASSR
         & \underline{37.56}, \underline{0.966}
         & 32.80, 0.898
         & 30.18, 0.807
         & N/A \\
        ARSSR
         & 35.58, 0.977
         & \underline{34.03}, \underline{0.959}
         & \underline{30.63}, \underline{0.938}
         & N/A \\
        \midrule
        \multicolumn{5}{c}{\textbf{Zero-shot methods}} \\
        \midrule
        NeRF
         & 33.96, 0.943
         & 27.03, 0.807
         & 29.71, 0.880
         & \underline{33.58}, \underline{0.938} \\
        CuNeRF
         & 31.03, 0.955
         & 28.36, 0.922
         & 27.12, 0.907
         & N/A \\
        Ours
         & \textbf{43.17}, \textbf{0.988}
         & \textbf{38.42}, \textbf{0.971}
         & \textbf{35.91}, \textbf{0.954}
         & \textbf{41.32}, \textbf{0.980} \\
        \bottomrule
    \end{tabular}}
\end{table}

\begin{table}[!t]
    \centering
    \caption{PSNR/SSIM for multi-scale SR on the FeTA dataset~\cite{xu2021feta}.}
    \label{tab:feta_results}
    \begin{tabular}{lccc}
        \toprule
        Method         & 2$\times$         & 3$\times$         & 4$\times$ \\
        \midrule
        %=== 第一组：常规模型 ===%
        \multicolumn{4}{c}{\textbf{Conventional methods}} \\
        \midrule
        Bicubic         & 37.29, 0.969         & 33.21, 0.913         & 29.62, 0.863 \\
        \midrule
        %=== 第二组：监督式方法 ===%
        \multicolumn{4}{c}{\textbf{Supervised methods}} \\
        \midrule
        MIASSR         & 38.71, 0.974         & 33.39, 0.921         & 30.10, 0.835 \\
        ARSSR         & \underline{40.99}, \underline{0.995}         & \underline{36.87}, \underline{0.988}         & \underline{34.45}, \underline{0.979} \\
        \midrule
        %=== 第三组：隐式零样本 NeRF-based 方法 ===%
        \multicolumn{4}{c}{\textbf{Zero-shot methods}} \\
        \midrule
        NeRF         & 40.96, 0.993         & 25.53, 0.859         & 32.43, 0.958 \\
        Ours         & \textbf{48.03}, \textbf{0.998}         & \textbf{43.52}, \textbf{0.995}         & \textbf{39.67}, \textbf{0.990} \\
        \bottomrule
    \end{tabular}
\end{table}

\begin{figure*}[!t]
  \centering
  \includegraphics[width=\linewidth]{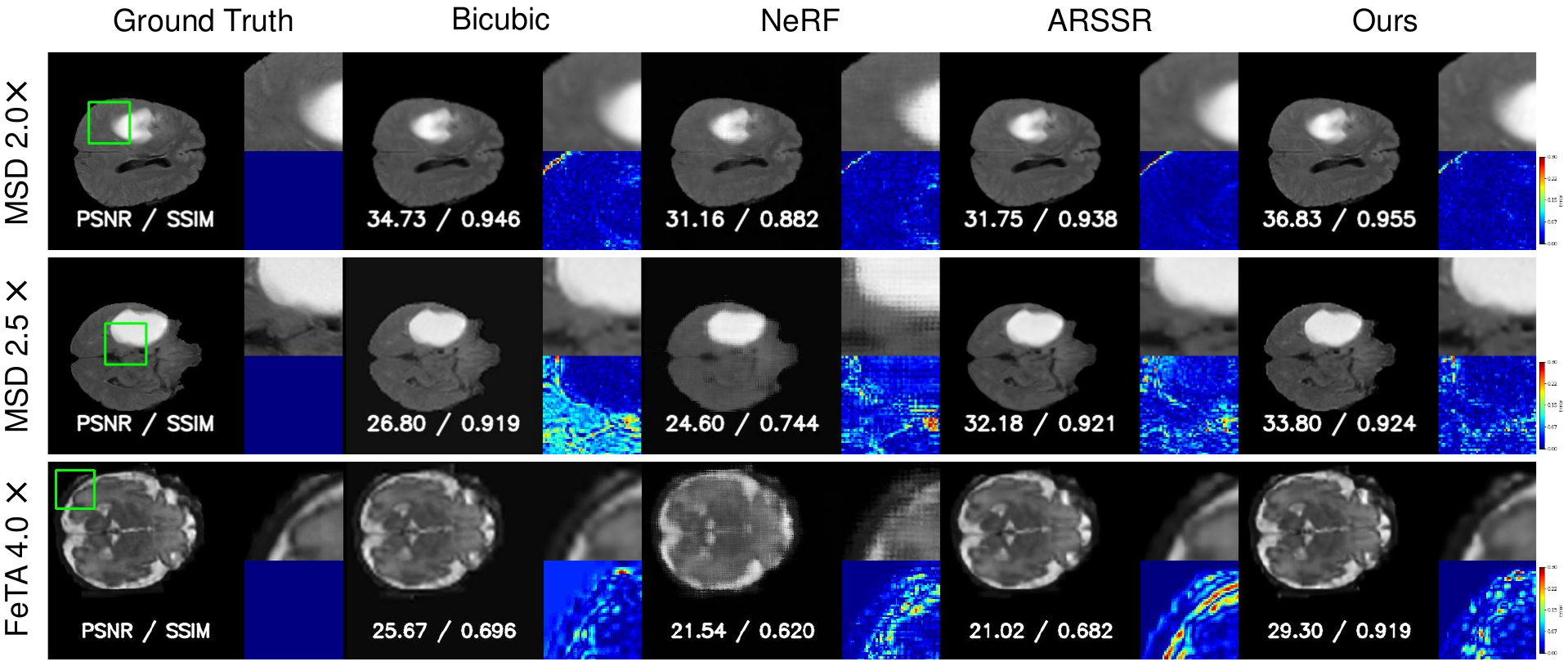}
\caption{
Visual comparison of the proposed method against state‑of‑the‑art approaches on two public datasets.
Left column: Full reconstructed images.
Top-right: Cropped region (indicated by a green box on the ground truth).
Bottom-right: Pixel-wise error maps between reconstructed patches and corresponding HR reference patches.
}
  \label{fig:main_result}
\end{figure*}

\section{Experiments}
\label{sec:experiments}

\subsection{Dataset}

We evaluate our method on two public 3D MRI datasets: the Brain Tumour dataset from the Medical Segmentation Decathlon (MSD)~\cite{simpson2019msd} (\(240\times240\times155\)) and the FeTA (Fetal Tissue Annotations) dataset~\cite{xu2021feta} (isotropic volumes, \(256\times256\times256\)).
 All images are intensity‑normalized to \([0,1]\) and downsampled via trilinear interpolation to generate low‑resolution inputs. For MSD, 100 subjects were randomly selected, of which 10 were held out for testing and the remaining 90 split into 80 for training and 10 for validation; zero‑shot models are trained only on the 10 test volumes (with HR data used solely for evaluation). For FeTA, 80 subjects were chosen, with 10 for testing and the other 70 split into 60 for training and 10 for validation; full volumes are used directly without patching.

\subsection{Comparison with Prior Work}

We compare our 3D Gaussian super-resolution framework with conventional interpolation, supervised CNN-based methods, and NeRF-style zero-shot approaches on the MSD and FeTA datasets. Reconstruction quality is evaluated using Peak Signal-to-Noise Ratio (PSNR) and Structural Similarity Index (SSIM) \cite{wang2004image}. As summarized in Tables~\ref{tab:brats_results} and \ref{tab:feta_results}, our method consistently achieves the highest PSNR and SSIM across scales and datasets. Figure~\ref{fig:main_result} visualizes the rendering results across different upsampling factors on both datasets.

Conventional interpolation such as Bicubic~\cite{keys1981cubic} yields the lowest reconstruction fidelity, while supervised CNN methods (e.g., ARSSR~\cite{zhang2021arssr}, MIASSR~\cite{yang2022miassr}) improve upon these results, but their performance still lags behind ours, especially at larger upsampling factors. Implicit zero-shot approaches (e.g., NeRF~\cite{mildenhall2020nerf}, CuNeRF~\cite{guo2023cunerf}) are attractive when paired LR–HR volumes are unavailable, yet in our experiments the explicit 3D Gaussian field better preserves anatomical detail and intensity characteristics, producing higher-fidelity reconstructions across a range of scales.

Beyond uniform upsampling,  which uniformly scales each axis by the same factor, our method can super-resolve to any user-specified target shape. For instance, on the MSD dataset, we start from a downsampled volume of size $100\times100\times100$ and recover the original resolution of $240\times240\times155$. By learning a global continuous spatial representation, we can sample at any user-specified voxel grid without retraining. This provides the flexibility of implicit continuous fields while retaining the fidelity and efficiency of an explicit, physics-grounded Gaussian model (see Table~\ref{tab:brats_results}).

\subsection{Ablation and Efficiency Analysis}

\begin{table}[!t]
    \centering
    % \scriptsize
    \caption{Ablation on amplitude ($A$) and relaxation proxy (T) for 3D MISR on the MSD dataset~\cite{simpson2019msd}. }
    \label{tab:ablation_halfcol}
    \renewcommand\arraystretch{1.1}
    \begin{tabular}{llccc}
        \toprule
        $a$ & rt & 2$\times$ & 3$\times$ & 4$\times$ \\
        \midrule
        $\times$     & $\times$ 
                    & 17.99, 0.825 
                    & 19.42, 0.817 
                    & 19.76, 0.790 \\
        $\times$     & $\checkmark$ 
                    & 34.56, 0.955 
                    & 29.83, 0.918 
                    & 27.77, 0.901 \\
        $\checkmark$ & $\checkmark$ 
                    & \textbf{43.03}, \textbf{0.988} 
                    & \textbf{38.57}, \textbf{0.972} 
                    & \textbf{36.15}, \textbf{0.957} \\
        \bottomrule
    \end{tabular}
\end{table}

\begin{figure}[!t]
  \centering
  \includegraphics[width=\linewidth]{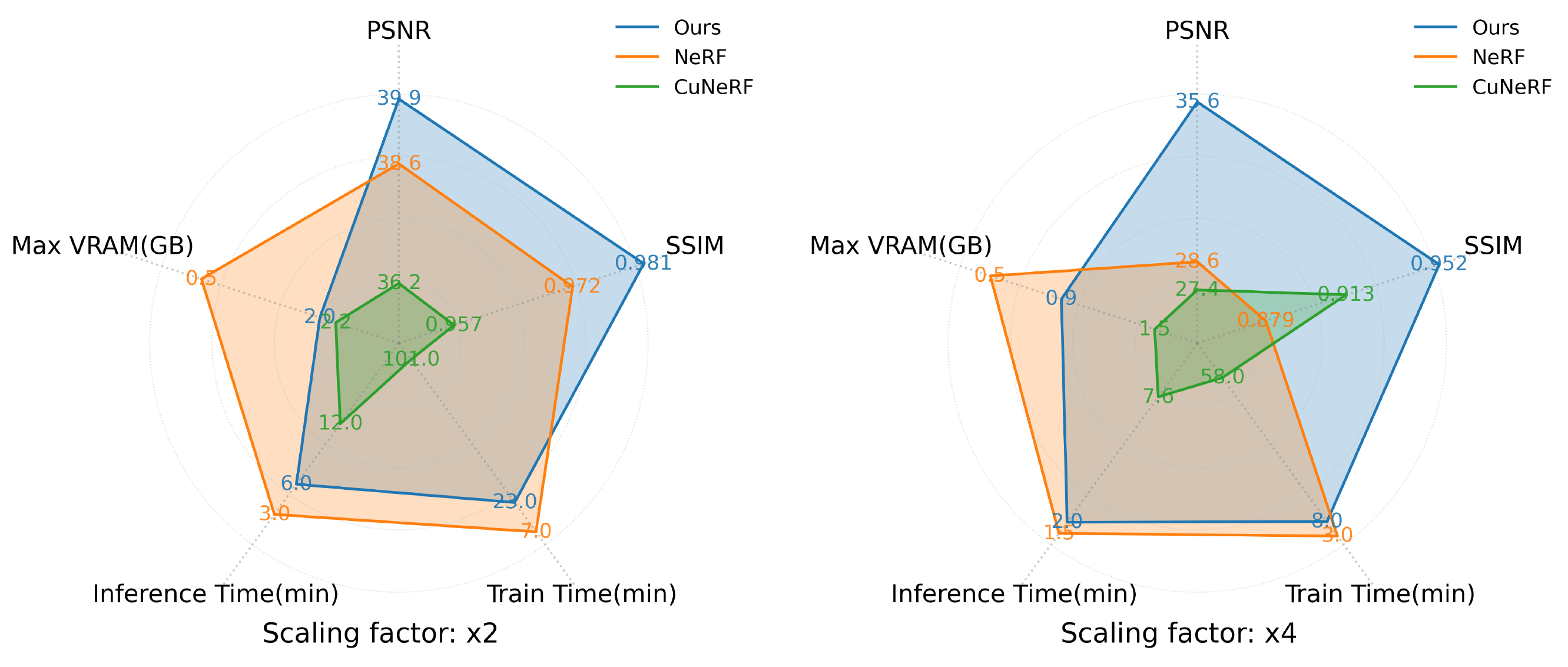}
  \caption{Radar chart of zero‑shot MRI SR methods (NeRF, CuNeRF, Ours) on MSD~\cite{simpson2019msd} at 2× (left) and 4× (right) upsampling. Metrics: PSNR/SSIM for quality and training time, inference time, peak VRAM for resources.}
  \label{fig:radar}
\end{figure}

We evaluate both the contributions of our Gaussian parameters and the efficiency of our approach. Ablation results (Table~\ref{tab:ablation_halfcol}) show that removing both amplitude and relaxation proxy yields nearly binary reconstructions lacking tissue structure, while adding only amplitude sharpens boundaries but misses fine textures. The full model recovers global structures and subtle details, demonstrating that each parameter meaningfully influences voxel reconstruction. For efficiency (Figure~\ref{fig:radar}), NeRF and CuNeRF either consume minimal memory but yield low PSNR/SSIM, or improve quality at the cost of long training and inference. Our method achieves the best balance, delivering high reconstruction quality with efficient training, inference, and memory usage. Notably, higher upsampling requires fewer Gaussians, reducing both computation and memory.

% \section{CONCLUSION}
% \label{sec:conclusion}
% In this work, we propose a physics-driven 3D Gaussian framework for zero-shot MRI super-resolution. Our method achieves high-quality reconstruction without paired data and with reduced computational cost, demonstrating strong potential for clinical applications.

\section{Conclusion}
\label{sec:conclusion}
To the best of our knowledge, this work presents the first explicit physics-driven 3D Gaussian point cloud framework for zero-shot MRI super-resolution, eliminating the need for paired LR–HR training data while achieving significantly reduced computational costs compared to implicit neural methods. Extensive evaluations on public MRI datasets demonstrate both superior image quality and computational efficiency, highlighting the promise of physics-driven models for clinical MRI super-resolution.

\bibliographystyle{IEEEbib}
\bibliography{strings,refs}

\end{document}